\documentclass[letterpaper, 10 pt, conference]{ieeeconf}  % Comment this line out if you need a4paper

\IEEEoverridecommandlockouts                              % This command is only needed if 
\usepackage{graphics} % for pdf, bitmapped graphics files
\usepackage{epsfig} % for postscript graphics files
\usepackage{amsmath} % assumes amsmath package installed
\usepackage{amssymb}  % assumes amsmath package installed
\usepackage{xcolor}

\DeclareMathOperator*{\argmax}{arg\,max}

\usepackage{fancyhdr}
\title{\LARGE \bf
A Model for Multi-Agent Autonomy That Uses Opinion Dynamics and Multi-Objective Behavior Optimization 
}

\author{Tyler M. Paine$^{1,2}$ and Michael R. Benjamin$^{1}$% <-this % stops a space
\thanks{$^{1}$Department of Mechanical Engineering, Massachusetts Institute of Technology, 
        Cambridge, MA 02139, USA
        {\tt\small tpaine@mit.edu, mikerb@mit.edu}}%
\thanks{$^{2}$Woods Hole Oceanographic Institution
        Woods Hole, MA 02543, USA}%
}

\begin{document}

\maketitle
\thispagestyle{empty}
\pagestyle{empty}

%%%%%%%%%%%%%%%%%%%%%%%%%%%%%%%%%%%%%%%%%%%%%%%%%%%%%%%%%%%%%%%%%%%%%%%%%%%%%%%%
\begin{abstract}
This paper reports a new hierarchical architecture for modeling autonomous multi-robot systems (MRSs):  a nonlinear dynamical opinion process is used to model high-level group choice, and multi-objective behavior optimization is used to model individual decisions. Using previously reported theoretical results, we show it is possible to design the behavior of the MRS by the selection of a relatively small set of parameters. The resulting behavior - both collective actions and individual actions - can be understood intuitively.  The approach is entirely decentralized and the communication cost scales by the number of group options, not agents.  We demonstrated the effectiveness of this approach using a hypothetical `explore-exploit-migrate' scenario in a two hour field demonstration with eight unmanned surface vessels (USVs). The results from our preliminary field experiment show the collective behavior is robust even with time-varying network topology and agent dropouts.  

\end{abstract}

\thispagestyle{fancy}
%%%%%%%%%%%%%%%%%%%%%%%%%%%%%%%%%%%%%%%%%%%%%%%%%%%%%%%%%%%%%%%%%%%%%%%%%%%%%%%%
\section{INTRODUCTION}

This paper reports a new hierarchical architecture for modeling multi-robot autonomous systems: opinion formation at the higher group-level, and multi-objective behavior optimization at the individual level.  Using previously reported theoretical results, we show the performance of the multi-robot system (MRS) can be designed by the selection of a relatively small set of parameters which reduces the complexity of performance optimization.  Additionally, inter-robot communication can be limited in range and have finite bandwidth which presents a challenge for the design of an autonomous MRS with a large number of robots. 
In this formulation, the group selects choices in a completely decentralized manner that is essential for scalability to larger group sizes.  

\begin{figure}
    \centering
    \includegraphics[width=1\columnwidth]{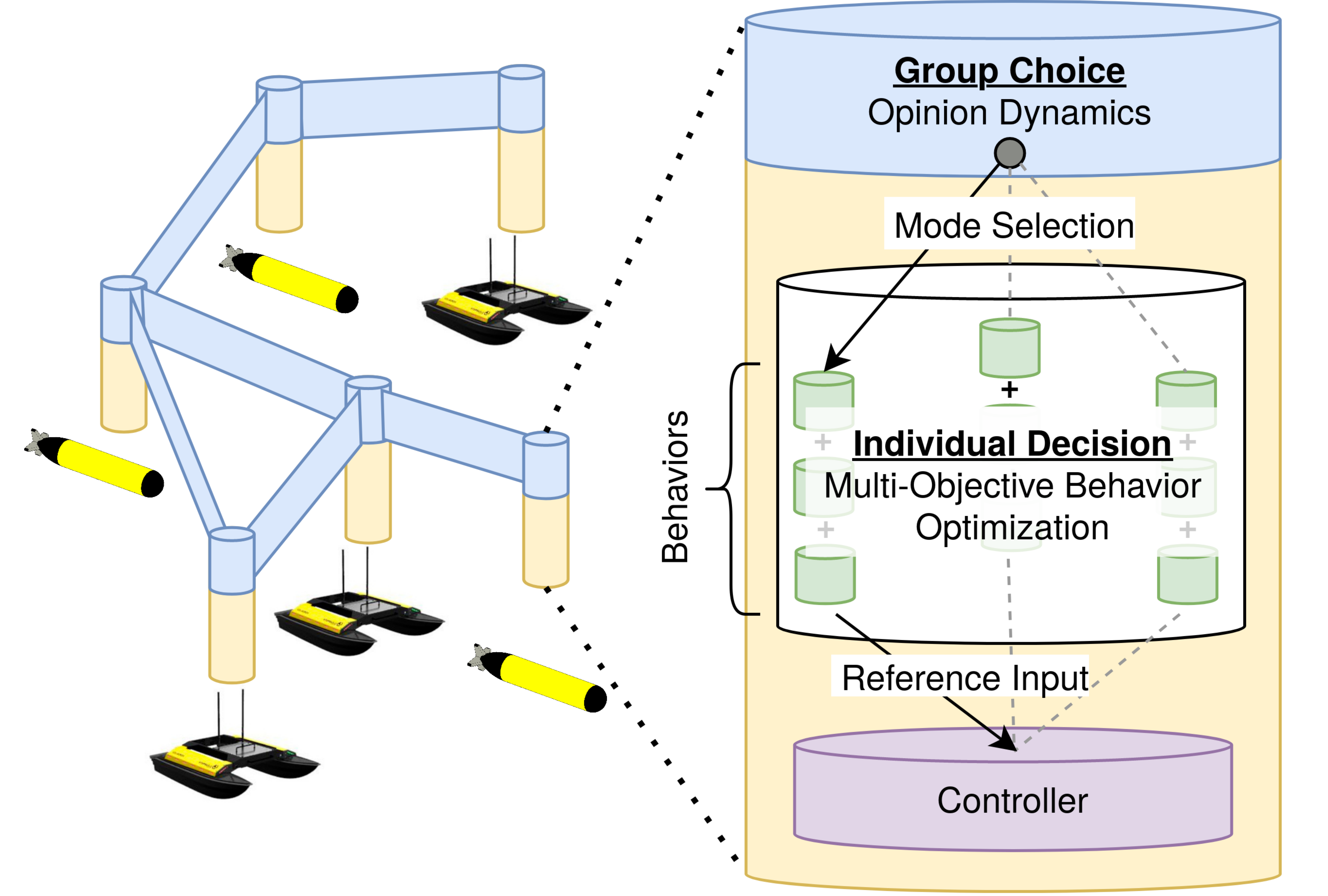}
    \caption{Overall framework of group choice with individual decision (GCID).  A networked system of robots is shown on the left, each sharing opinions with neighbors (blue).  The group choice selects which local behaviors are active and the behavior optimization decides the reference input to the lower-level controller (yellow).}
    \label{fig:enter-label}
    \vspace{-5mm}
\end{figure}
This framework for autonomous decision-making includes two major levels: group choice that occurs among networked robots, and individual decisions that are made locally.   In this work we model and parameterize group choice as a nonlinear dynamical opinion process  \cite{Bizyaeva2023OD_TAC}.  Recent research has shown that this model of opinion dynamics is able to capture a wide range of functionality required of large networked systems including the ability of a group to achieve consensus and dissensus, break deadlock, and cascade important opinions across a group \cite{franci2021analysis}. At the individual level, robots select values of heading and speed - or any other decision variable - that maximizes utility in the sense of multi-objective behavior optimization \cite{Benjamin2010MOOS}.  
This optimization process generates a desired trajectory, or a reference input.  Although not discussed in detail in this paper, a controller is used to drive the actual state of the vehicle along the desired trajectory. 

Contributions described in this paper include:
\begin{enumerate}
    \item A new architecture, group choice with individual decision (GCID), for explainable decentralized multi-agent autonomy that combines a group-level choice modeled as a dynamical system of opinions with an individual-level decision making via interval programming (IvP)
    \item A two hour field demonstration of this approach using a fleet of eight unmanned surface vessels (USVs) in an explore-exploit-migrate scenario, along with Monte-Carlo simulation studies. 
\end{enumerate}

\section{RELATED WORK}
It is widely accepted that networked multi-robot systems are more capable of completing tasks more efficiently as well as being more robust to failures \cite{rizk2019cooperative}, \cite{verma2021multi}. However, one essential prerequisite is that the autonomous population must exhibit collective intelligence \cite{Leonard2022PublicGoods}.  A group has collective intelligence if it is able to complete problems such as task allocation \cite{Gershfeld2023RAL} \cite{Park2021ICRA}, adaptive sampling \cite{Leonard2007IEEE} \cite{Kemna2017ICRA}, and formation assembly \cite{Benjamin2021Oceans}, \cite{REN200RAS} in response to environmental stimuli.  Collective intelligence in multi-robot systems is typically designed using algorithms that exploit shared information, but can also emerge from the aggregation of simple actions from individuals with only a limited understanding of their immediate surroundings \cite{Rubenstein2014Science}.  In general, more sophisticated group behaviors are associated with greater communication between individuals in combination with increased sensory feedback and better reasoning or planning at the individual level \cite{rizk2019cooperative}, \cite{verma2021multi}.  

Collective behaviors of a population are a result of both competitive and cooperative behaviors by individuals \cite{verma2021multi}, \cite{Leonard2022PublicGoods}.  The authors recently fielded a MRS which provides a clear example of this insight where a group of USVs cooperatively found deep channels in rivers \cite{Gershfeld2023RAL}.  In that work, individual robots cooperated in building a map of the river using a consensus protocol, but then competed against each other for the most promising routes to explore using a market-based proposal bidding scheme.  Complex collective behaviors present in biological groups of animals emerge from the opposing incentives to cooperate for safety yet compete for resources \cite{couzin2009collective}. However, some MRS are limited because they are programmed to perform only competitive behaviors such as auctioning algorithms \cite{Benjamin2021Oceans} or only cooperative behaviors such as adaptive sampling \cite{Kemna2017ICRA}.  In this work, we aim to formulate a general model of robotic populations that is able to express a wide range of collective behaviors by including both competition and cooperation amongst individuals. 

The combination of competition and cooperation behavior in MRS is necessary for coalition formation, where sub-groups of robots choose to work together amongst themselves, but not necessarily with other robots outside the sub-group.  Coalition formation can be thought of as either separate from task allocation or as a process that occurs simultaneously with task allocation \cite{rizk2019cooperative}, e.g. consensus-based bundle adjustment (CBBA) \cite{Luc2009CBBA}.   There are many general approaches to solving a task allocation problem such as precisely encoding the tasks and holding an iterative auction \cite{Luc2009CBBA}, writing time and space constraints and solving a temporal logic optimization problem for individual robot policies as in  \cite{Leahy2022ScRATCHes}, \cite{liu2023catlnet}, and defining one or more reward functions and using reinforcement learning to learn policies \cite{tian2009multi}.  Clearly, once a robot makes the choice to join a group there are many ways to collaborate.  However, in this work we are interested in defining a model for dynamic coalition formation that can accommodate any style of task allocation and where coalitions perpetually reform throughout the life of the autonomous population. 

Decentralized approaches to coalition formation are in general preferred because such systems are robust to  communication drop-outs and failures of individual robots which happen periodically in the field.  Decentralized Monte-Carlo tree search (MCTS) has been used for coalition formation as reported in \cite{Zheng2002MCTS}, although most applications use MCTS for lower level path planning \cite{Karl2018MCTS}, \cite{Daneshvaramoli2020ComMCTS}.  However, in general the communication bandwidth required for MCTS approaches scales linearly by the number of agents (and the size of their state), whereas the communication cost for the opinion dynamics model used in this study scales linearly by the number of options, not agents.   This property of the opinion dynamics model is increasingly important as the population grows in size.  Finally, although not discussed in detail in this paper, MCTS methods theoretically require far greater computational resources than this new GCID approach. 

\section{NOTATION AND DEFINITIONS}
We denote vectors as $\vec{v} \in {\rm I\!R}^n$. The $n$-dimensional vector of all ones is $\vec{1}_n$.    We define the strictly ordered set of mutually orthogonal standard unit vectors in $n$-dimensional space with $n\geq 2$ as $E(n,<)$ where 
\begin{align}
E(n, <) = \big\{ \vec{e_i} :  &\text{ where \text{$e_i < e_j$ with $i<j$} \big\}} \nonumber
\end{align}

The state for the $i^{th}$ agent is $\vec{x}_i \in {\rm I\!R}^n$ which includes kinematic states (pose and velocities) in addition to other states such as fuel level.  The state vector $\vec{x}$ importantly does not contain information about the \emph{opinion state} of the agent which is defined in Section \ref{sec:op_dynam}.  
The connections between agents in networked systems is represented by an unweighted adjacency matrix $\bar{A}$ with entries $\bar{a}_{ik}$ that includes self-loops.  $\bar{a}_{ik} = 1$ if the $i^{th}$ agent in connected to the $k^{th}$ agent and $\bar{a}_{ik} = 0$ otherwise. The set of agents connected to the $i^{th}$ agent is $\Xi_i$

\section{GROUP CHOICE VIA OPINION DYNAMICS} \label{sec:op_dynam}
In the group choice individual decision (GCID) framework, group choice is modeled as a dynamical system of opinions.  The opinion formation process is decentralized and robust to time-varying changes in network topology.  Each agent holds an opinion of $N_O$ options using the following notation adopted from the work of Bizyaeva \textit{et al.} \cite{Bizyaeva2023OD_TAC}:  Opinions of the $i^{th}$ agent are modeled as a vector $\vec{z}_i$, where $z_{ij}$ is the opinion of the $i^{th}$ agent about the $j^{th}$ option, and the opinions of each agent sum to zero.  More formally, $\vec{z} \in \vec{1}\frac{1}{N_O} \subset  {\rm I\!R}^{N_O}$. The projection onto $\vec{1}\frac{1}{N_O}$ is defined as $P_0 = I_{N_O} - \frac{1}{N_O} \vec{1}_{N_O} \vec{1}^T_{N_O}$. As defined in \cite{Bizyaeva2023OD_TAC} positive values for $z_{ij}$ imply the $i^{th}$ vehicle has a positive opinion about the $j^{th}$ option.  

We use the Heterogeneous Inter-option Coupling model from \cite{Bizyaeva2023OD_TAC} for a population of $N_a$ agents:
\begin{align}
\dot{\vec{z}}_i =& P_0 F_i(\vec{z}_i), \\
F_{ij} =& -d_i z_{ij} + u_i \sum_{l = 1}^{N_O} S \bigg(\sum_{k = 1}^{N_a} A_{ik}^{jl} z_{kj} \bigg) + b_{ij}, \label{eq:op_dynam1}
\end{align}
where the adjacency tensor is denoted as $A\in {\rm I\!R}^{N_a \times N_a \times N_O \times N_O}$ with entries $A_{ik}^{jl}$ that parameterize the influence from the $k^{th}$ agent's opinion about option $l$ on the $i^{th}$ agent's opinion about option $j$.  We have an intuitive understanding of the parameters in this model from previous theoretical work \cite{Bizyaeva2023OD_TAC}:
\begin{itemize}
\item $A_{ii}^{jj}$ is the intra-agent, same-option coupling.  We restrict $A_{ii}^{jj} \geq 0$, and if $A_{ii}^{jj} > 0$ the agent's opinion about this option is self-reinforcing.  
\item $A_{ii}^{jl}$ is the intra-agent, different-option coupling.  This parameter is used to encode interplay between agent's own opinions.
\item $A_{ik}^{jj}$ is the inter-agent, same-option coupling.
\item $A_{ik}^{jl}$ is the inter-agent, different-option coupling.
\end{itemize}
In homogeneous systems where $A_{ik}^{jl_1} = A_{ik}^{jl_2} \ \ \forall \ l_1, l_2 = 1, 2, \hdots N_O \ l_1, l_2 \neq j$ and $A_{ik}^{j_1j_1} = A_{ik}^{j_2j_2} \ \ \forall \ j_1, j_2 = 1, 2, \hdots N_O$, cooperation and competition between agents are parameterized by $A_{ik}^{jj}$ and $A_{ik}^{jl}$ as: \cite{Bizyaeva2023OD_TAC}
\begin{align}
A_{ik}^{jj} - A_{ik}^{jl} >& \ 0 \  \rightarrow \ \text{Cooperation} \\
A_{ik}^{jj} - A_{ik}^{jl} <& \ 0 \  \rightarrow \ \text{Competition} 
\end{align}
%As explained in detail in \cite{Bizyaeva2023OD_TAC}, if the net effect of the influence is negative then agent will tend to form the opposite opinion of its neighbors which is competitive behavior. 
The parameter $d_i > 0$ and $u_i$ in  (\ref{eq:op_dynam1}) are the resistance of the opinion and the tunable attention parameter, respectively.

We find in both simulation and field experiments that tunable attention with saturation is particularly useful when designing group behavior that must transition from a mode of normal operation to one of emergency that requires a heightened sense of urgency.  Attention for the $i^{th}$ agent is modeled as a saturating function of the magnitude of all neighbor's opinions, 
\begin{align}
\tau_u \dot{u}_i = -u_i + S_u\bigg( \frac{1}{N_O} \sum_{k=1}^{N_a} \sum_{l=1}^{N_O} \big( \bar{a}_{ik} z_{kl} \big)^2 \bigg), \label{eq:atten_dynam}
\end{align}
where $\tau_u$ is the time constant and $S_u$ is the Hill saturating function with parameters as described in \cite{Bizyaeva2023OD_TAC}. 

Since this formulation is completely decentralized, the input $b_{ij}$ for each opinion is computed using only locally known information.  This knowledge includes the state of the agent and any information shared between adjacent agents.  However, to prevent redundancy with the parameters entering into the dynamical systems (\ref{eq:op_dynam1}) and (\ref{eq:atten_dynam}) the input $b_{ij}$ is defined to be a utility function of only the states $\vec{x}$
\begin{equation}
b_{ij} = g_j(\vec{x}_i, \vec{x}_k)  \ \forall \ k \in \Xi_i.  \label{eq:group_utility}
\end{equation}

Since each agent is constantly evaluating their own inputs and the opinions of neighbors, the choice at the group-level is continuously changing throughout the lifespan of the population.  To connect the group-level choice to agent-level decision-making, we define $\vec{e}_{z_{max}}$ to represent the most favorable ordered option as a unit vector by
\begin{align}
\vec{e}_{z_{max}} = \min_{E} \ \big( \text{arg} \max_{\vec{e}_i \in E} (\vec{e}^T_i \vec{z}_i ) \big), \label{eq:z_max_def}
\end{align}
e.g for $z_2 = z_3 > z_j \ \forall \ j \neq 2,3$, then
\begin{align}
 \vec{e}_{z_{max}} =& \begin{vmatrix}
 0 &
 1 &
 0 &
 \hdots &
 0
 \end{vmatrix}^T.
\end{align}

\section{INDIVIDUAL DECISION-MAKING VIA IvP}
In the group choice individual decision (GCID) framework, individual decision-making is modeled as an interval programming (IvP) optimization problem \cite{benjamin2004interval}.  The optimizer computes the pareto optimal solution for the reference trajectory given the objective functions of all active behaviors.  However, in this framework, the process of group choice described in Section \ref{sec:op_dynam} determines for each agent which behaviors are active.  The remainder of this section provides a description of the IvP process within the context of the GCID framework.  

Usually, a single behavior is programmed for an individual agent to achieve a singular desired outcome, although some behaviors are designed to participate in a structured task allocation problem with neighboring agents \cite{Benjamin2021Oceans} \cite{Gershfeld2023RAL}.  A behavior maps the values of decision variables, such as desired heading and desired speed, to a value of utility. 
The decision space, $S_m$, for each of $m$ decision variables, $r_m$, is assumed to be finite and uniformly discrete, i.e. $r_m \in S_m \subset {\rm I\!R}$. 
The $q^{th}$ behavior generates an objective function $f_{q}(r_1, r_2, \hdots r_m): (S_1 \times S_2 \times \hdots S_m) \rightarrow {\rm I\!R}$.  
More complicated autonomy can be expressed through a combination of several behaviors, and IvP is used to balance competing objectives. 

The multi-objective optimization problem can be written as \cite{Benjamin2010MOOS}
\begin{align}
r_1^*, r_2^*, \hdots r_m^* = \argmax_{r_m \in S_m \ \forall  m} \sum_{q=1}^{N_{active}} w_{q} f_{q}(r_1, r_2, \hdots r_m)
\end{align}
where the utility function of the $q^{th}$ behavior is weighted by $w_{q} \in {\rm I\!R}$ and $N_{active}$ is the number of active behaviors.  Typically, the pareto optimal decisions of $r_1^*, r_2^*, \hdots r_m^*$ are included in the reference signal $\vec{r}(t)$ sent the vehicle controller.

In this new framework, the strongest opinion as calculated in (\ref{eq:z_max_def}) selects the behaviors that are active. 
\begin{align}
r_1^*, r_2^*, \hdots r_m^* = \argmax_{r_m \in S_m \ \forall  m} \vec{e}_{z_{max}}^T 
A_{c} W
\begin{bmatrix}
f_{1}(r_1, r_2, \hdots r_m) \\
f_{2}(r_1, r_2, \hdots r_m) \\
\vdots \\
f_{q}(r_1, r_2, \hdots r_m)
\end{bmatrix},
\end{align}
where the diagonal weighting matrix 
\begin{equation}
 W = diag( \begin{bmatrix}
w_1 & w_2 & \hdots & w_q
\end{bmatrix}). 
\end{equation}
The matrix $A_{c}$ is the mapping from group-choice options options to active behaviors on individual agents.  All entries $A_{c_{jq}} \in \{0,1\}$ are freely chosen design parameters.   

In this formulation, the typical rule-based selection of active behaviors - known as MODE selection in MOOS-IvP - is replaced with a dynamic process of activating behaviors that are associated with the strongest positive opinion of the agent.  As discussed in Section \ref{sec:op_dynam}, the outcome of this process depends not only upon the local utility of each option (\ref{eq:group_utility}), but also upon the opinions of immediate neighbors. 

\section{IMPLEMENTATION DETAILS} \label{sec:impl_details}
The GCID framework requires two major components: the opinion formation system and the multi-objective behavior optimization via IvP.  

\subsection{Opinion Manager Engine}
\begin{figure}
    \centering
    \includegraphics[width=1\columnwidth]{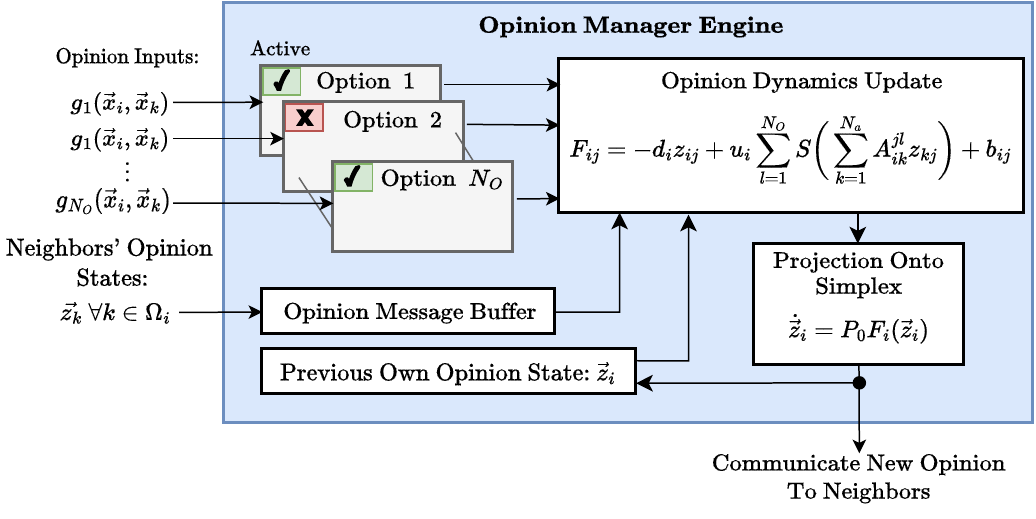}
    \caption{Overview of Opinion Manager Engine which separately runs on each vehicle in a decentralized scheme. At each iteration, local opinion inputs, any available opinion of neighbors, and previous own opinion state are used to determine the next opinion state.  The new state is communicated to any nearby vehicles. }
    \label{fig:op_manager_engine_overview}
    \vspace{-5mm}
\end{figure}

We wrote the ``Opinion Manager Engine'' to manage the dynamical process of opinion formation in a completely decentralized manner that is robust to changing network topology and communication dropouts.  An overview of the Opinion Manager Engine is provided in Figure \ref{fig:op_manager_engine_overview}. 

On startup, the Opinion Manager Engine loads a configuration file that specified the properties of any number of possible options to be considered.  All the option parameters related to coupling, activation, and input variables are exposed as configuration parameters to eliminate changes to the core code of the Opinion Manager Engine.  This design choice allows the same Opinion Manager Engine to be used for any multi-vehicle mission, provided the user specifies the options with the desired parameters. 

The Opinion Manager Engine handles all asynchronous communication between vehicles using a message buffer protocol where outdated opinions are ignored. 
Parameters related to the dynamics of option formation, such as those in (\ref{eq:op_dynam1}) and (\ref{eq:atten_dynam}), are specified as configuration parameters for the Opinion Manager Engine itself.  The iteration frequency of Opinion Manager Engine can also be specified, and for our field testing we found a frequency of 1 Hz provided adequate response time with a relatively low bandwidth of 3-12 bytes per second depending on resolution.  A MOOS App wrapper was used to connect the Opinion Manager Engine to other apps using the MOOS middleware.

\subsection{Interval Programming (IvP) \cite{Benjamin2010MOOS}}
Individual behavior optimization was performed using the MOOS-IvP Helm, which is well documented in \cite{Benjamin2010MOOS} and \cite{benjamin2010UserMan}.  However, in this work we departed from the typical convention of hierarchical mode selection where the designer prescribes every mode via a collection of logic statements.  Instead, we configured the behavior file so that some modes are selected by a rigid set of conditional statements, for example modes related to safety such as \texttt{RETURN} or \texttt{ALL\_STOP}, while other modes could be chosen by the dominant opinion of the individual agent - the output of the Opinion Manager Engine shown in Figure \ref{fig:op_manager_engine_overview}.  In the field experiments with USVs, the decision variables were desired speed and desired heading.

\section{EXPERIMENTAL SETUP} \label{sec:field_exp}
We demonstrated the effectiveness of this approach using a hypothetical `explore-exploit-migrate' scenario that is cast as the problem of detecting and sampling algae blooms using a persistently deployed autonomous population of robotic vehicles.   At the core of many scenarios in multi-agent autonomy is a tension between allocating resources to explore or to exploit.  In this work we added a third option, known as migration, where a group comes to a consensus about moving to a new region and completes the move together.  In this specific bloom detection scenario, the three options are described as follows:
\begin{itemize}
    \item \textbf{Explore:} Vehicles search the region to find indication that an algae bloom was present.  In this scenario vehicles are equipped with a sensor to measure turbidity in the water - since high levels are a good predictor of an algae bloom \cite{rome2021sensor}  - but the vehicles must move slowly for sensor readings to be accurate.  
    \item \textbf{Exploit:} Vehicles quickly move to take samples at locations where algae blooms are indicated by vehicles that are exploring.  Vehicles must stop to take a sample, which hypothetically would be transported to a lab for confirmation.  
    \item \textbf{Migrate:} Vehicles quickly move together to a new region, usually after detecting a source of danger - such as a storm - or lack of algae blooms to sample. 
    
\end{itemize}

\begin{figure*}
    \centering
    \includegraphics[width=1\textwidth]{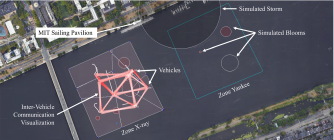}
    \caption{State of the MRS and simulated environment approximately 50 minutes into a 2 hour mission on the Charles River near the MIT Sailing Pavilion.  Simulated blooms appeared randomly within both zones X-ray and Yankee, and grew larger with time.  A randomly generated storm periodically passed over the regions. At this time during the mission, 8 vehicles were searching and sampling in Zone X-ray.  The communication range was artificially limited to 160 meters, and the restricted inter-vehicle communication is visualized in red. Both zones measured 300 meters by 350 meters.   }
    \label{fig:BloomStormMission}
    \vspace{-5mm}
\end{figure*}

An overview of the scenario is shown in Figure \ref{fig:BloomStormMission}.  Simulated harmful algae blooms were generated using a environmental simulator where the bloom area starts from a randomly generated location within the zone and grows outward for 10 minutes. The primary goal in this scenario was for the MRS to find the blooms and sample nearby water for confirmation.  A simulated storm was randomly generated to pass over one zone, and a secondary design goal of the MRS was to avoid the storm as much as possible by moving to the other zone when a storm is detected.

\subsection{Heron USVs}
\begin{figure}
    \centering
    \includegraphics[width=1\columnwidth]{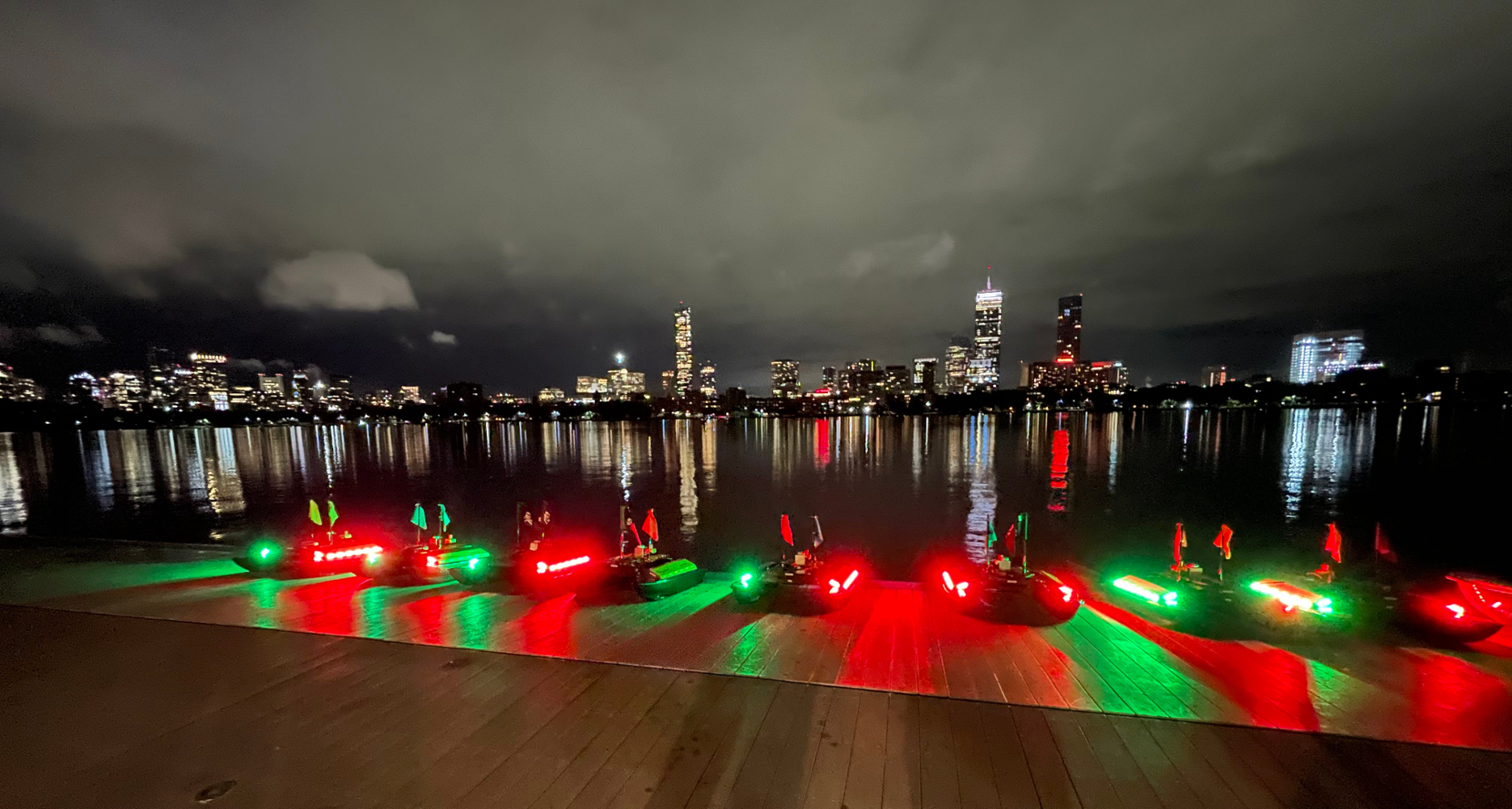}
    \caption{8 Heron USVs used in a two-hour field deployment at night on the Charles River in Boston, MA }
    \label{fig:herons_night}
   \vspace{-5mm}
\end{figure}

In the field experiments we used a fleet of eight Heron USVs made by Clearpath Robotics as shown in Figure \ref{fig:herons_night}.  These USVs are designed using the frontseat-backseat paradigm, and the autonomy stack was loaded on the backseat computer.   Each vehicle is equipped with a GPS and IMU sensor used for state estimation, and have wireless communication with a router at the MIT sailing pavilion. The vehicle names are: Abe, Ben, Deb, Eve, Fin, Max, Ned, and Oak.

\subsection{Autonomy Configuration}
We enumerated the options as $1: \texttt{Explore}$, $2: \texttt{Exploit}$, $3: \texttt{Migrate}$.  
The relationship between key parameters in the heterogeneous inter-option coupling model  (\ref{eq:op_dynam1}) were:
\begin{align}
    A_{ik}^{jj} - A_{ik}^{jl} & > 0, \ j = 3, \ \forall l, \\
    A_{ik}^{jj} - A_{ik}^{jl} & < 0, \ j = 1,2, \ l = 1,2,  \\
    A_{ik}^{jj} - A_{ik}^{jl} & > 0, \ j = 1,2, \ l = 3. 
\end{align}
The design intent was to have the explore-exploit options selected primarily on a basis of social competition, while the migrate option was selected cooperative via an opinion cascade. The attention system (\ref{eq:atten_dynam}) was tuned so that the social influence increased when the magnitude of the opinions increased due to a storm detection.  The input $b_{ij}$ for each option was designed as follows:
\begin{itemize}
    \item Explore: $g_1(\vec{x}_i, \vec{x}_k)$ increases linearly with distance traveled (in simulation) or with battery level consumed (in field experiments), in addition to a positive bias when no samples are available. 
    \item Exploit: $g_2(\vec{x}_i, \vec{x}_k)$ increases linearly with range to nearest available sample with a maximum value, in addition to a positive bias when currently sampling. 
    \item Migrate: $g_3(\vec{x}_i, \vec{x}_k)$ includes a large positive bias when a storm is detected, in addition to a positive bias when migrating and all neighbors are not yet within the destination region.  A negative bias is included after a migration is complete and it increases linearly with time since the last migration until it is equal to zero.
\end{itemize}
Due to space constraints we do not list the specific values, but they are available upon request. 

Finally, since at any given time there were multiple vehicles exploring the same region and multiple vehicles traveling to multiple sample locations, two different schemes for task allocation were required.  Exploring vehicles used a search algorithm based on Voronoi partitioning previously published in \cite{evans2022practical}.  For sampling vehicles, an optimal route for sampling vehicles can be found by solving a distributed multiple traveling salesman problem - which is at least NP-complete \cite{russell2010artificial}.  In this work sampling vehicles used a greedy approach where they traveled to the closest point which was not also the closest point to another sampling vehicle. 

\section{SIMULATION RESULTS}
\begin{figure}
    \centering
    \includegraphics[width=1\columnwidth]{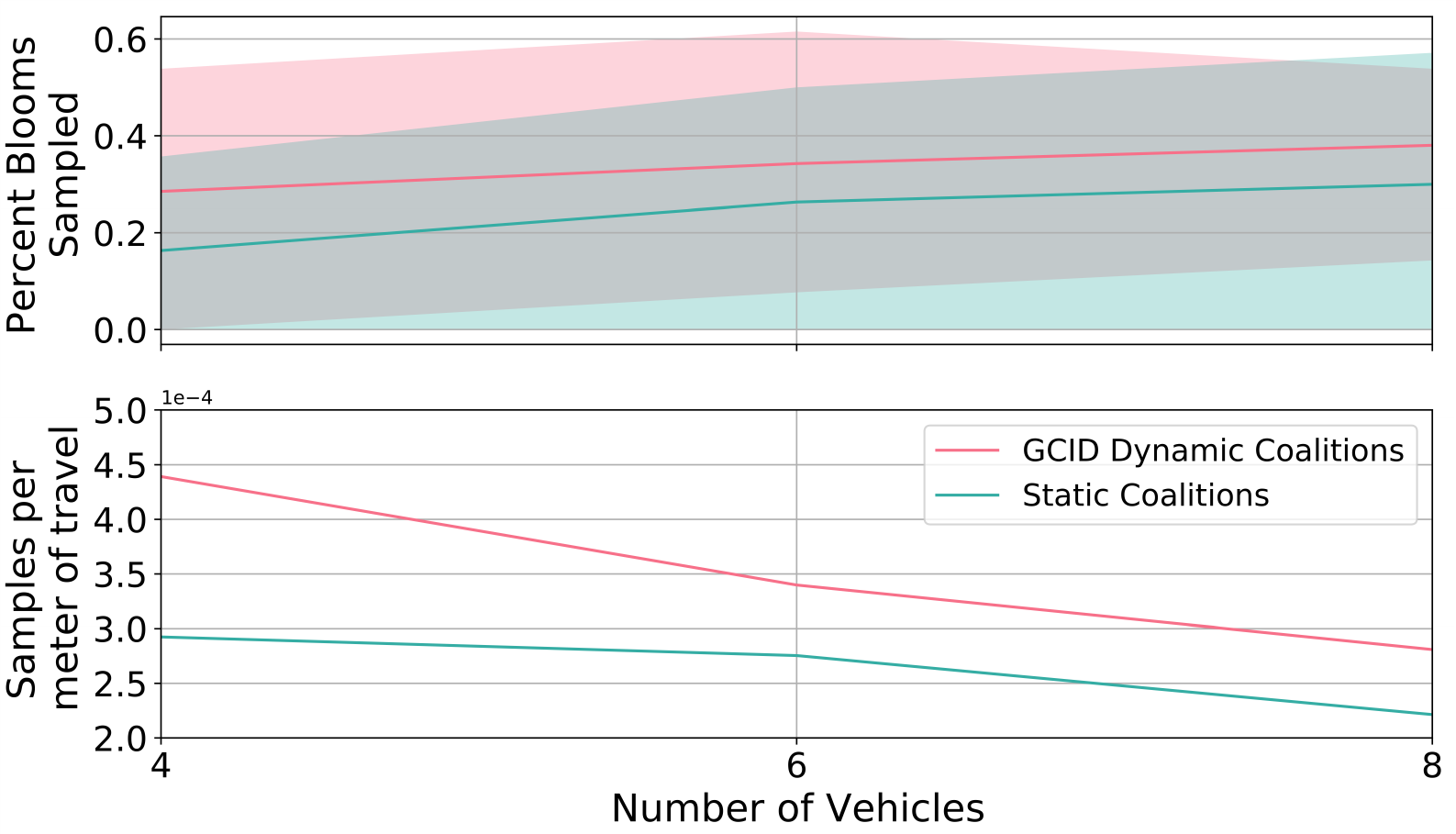}
    \caption{Monte Carlo simulation experiments of the bloom sampling scenario with increasing fleet size.  The MRS using the GCID approach sampled a higher percentage of blooms on average (top), and also sampled blooms with a slightly higher average efficiency - greater samples per meter of travel (bottom). Solid lines are the mean for each type and the shaded areas indicate the min and max values of all trials.}
    \label{fig:MC_Sim_fig}
    \vspace{-5mm}
\end{figure}
Due to the limited amount of uninterrupted time available on the Charles River as well as the onerous supervision required for a large MRS that operated in 0.21 $km^2$ of busy public water-space, we conducted a series of preliminary Monte-Carlo simulation studies to evaluate the effectiveness of this method.  For comparison we fixed the coalitions in the MRS such that in the absence of a storm half the fleet was exploring, and the other half exploiting.  This composition of modes in the static coalition matched the collective behavior designed using the GCID framework.  Simulations were completed with fleets of size 4, 6, and 8.  A total of 300 simulations were completed using Monte-MOOS \cite{Becker2024Thesis}, each with randomized blooms and storms. 

A summary of results is shown in Figure \ref{fig:MC_Sim_fig}. On average the GCID dynamic coalition formation outperformed the static coalitions on the basis of sampling percentage and efficiency. The same task allocation algorithms were used in both cases, and the only difference was in allowing group choice via opinion dynamics.

\section{FIELD RESULTS}
\begin{figure}
    \centering
    \includegraphics[width=1\columnwidth]{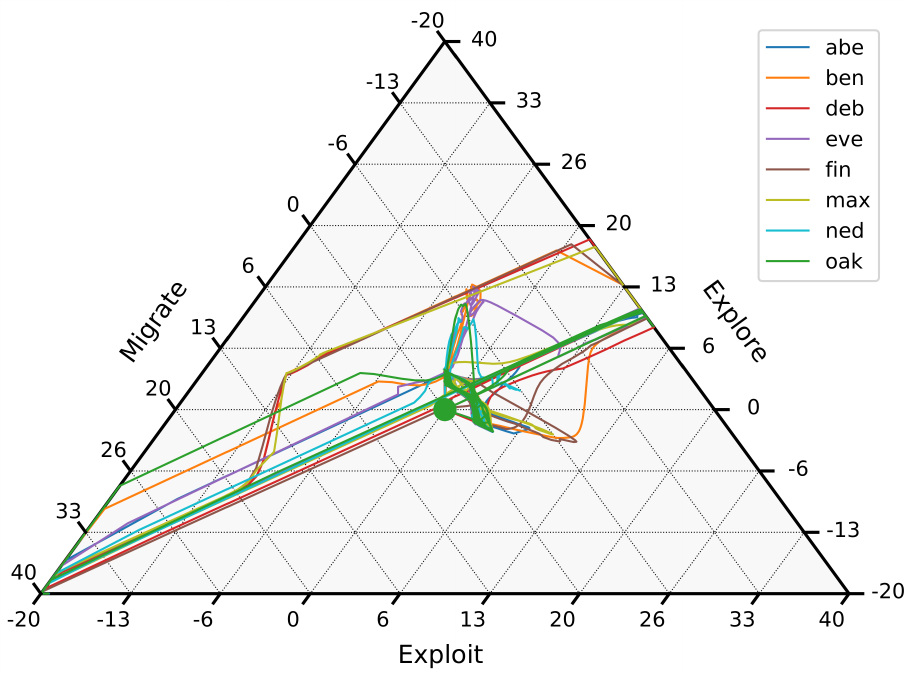}
    \caption{ Trajectories of the opinion state for each vehicle during the 2 hour field mission.   Each trace shows how the opinion for a vehicle evolved over time in response to input and opinions from neighbors.  This detail view on the simplex is zoomed in to the range [-20, 40] to show the trajectories near the origin. }
    \label{fig:opinion_traject}
    \vspace{-5mm}
\end{figure}

We report field results from a 2 hour night-time operation on the Charles River.  Using the GCID approach, all 8 USVs dynamically changed coalitions, locating possible simulated blooms and allocating vehicles to sample them.  All vehicles selected to start in Zone X-ray and all migrated to Zone Yankee when one vehicle (Abe) detected a storm.  A plot of the opinion trajectories is given in Figure \ref{fig:opinion_traject} which shows two interesting features: \textbf{1)} A clear bias toward exploring during portions of the mission when there are no known locations to sample, and \textbf{2)} a bifurcation near the origin between the options to explore and exploit, with evidence that vehicle's opinion state would transition between the two options as desired.

A plot of the magnitude of the attention (\ref{eq:atten_dynam}) for each vehicle is shown in the top plot of Figure \ref{fig:atten_deg}.  As designed, the periods of high attention correspond to either participating in a migration or just completing a migration.  In this way, attention was tuned such that the MRS was responsive to the detection of a storm by any one vehicle, and that response was an opinion cascade for every vehicle in the network. 

\subsection{Robustness to vehicle dropout}
\begin{figure}
    \centering
    \includegraphics[width=1\columnwidth]{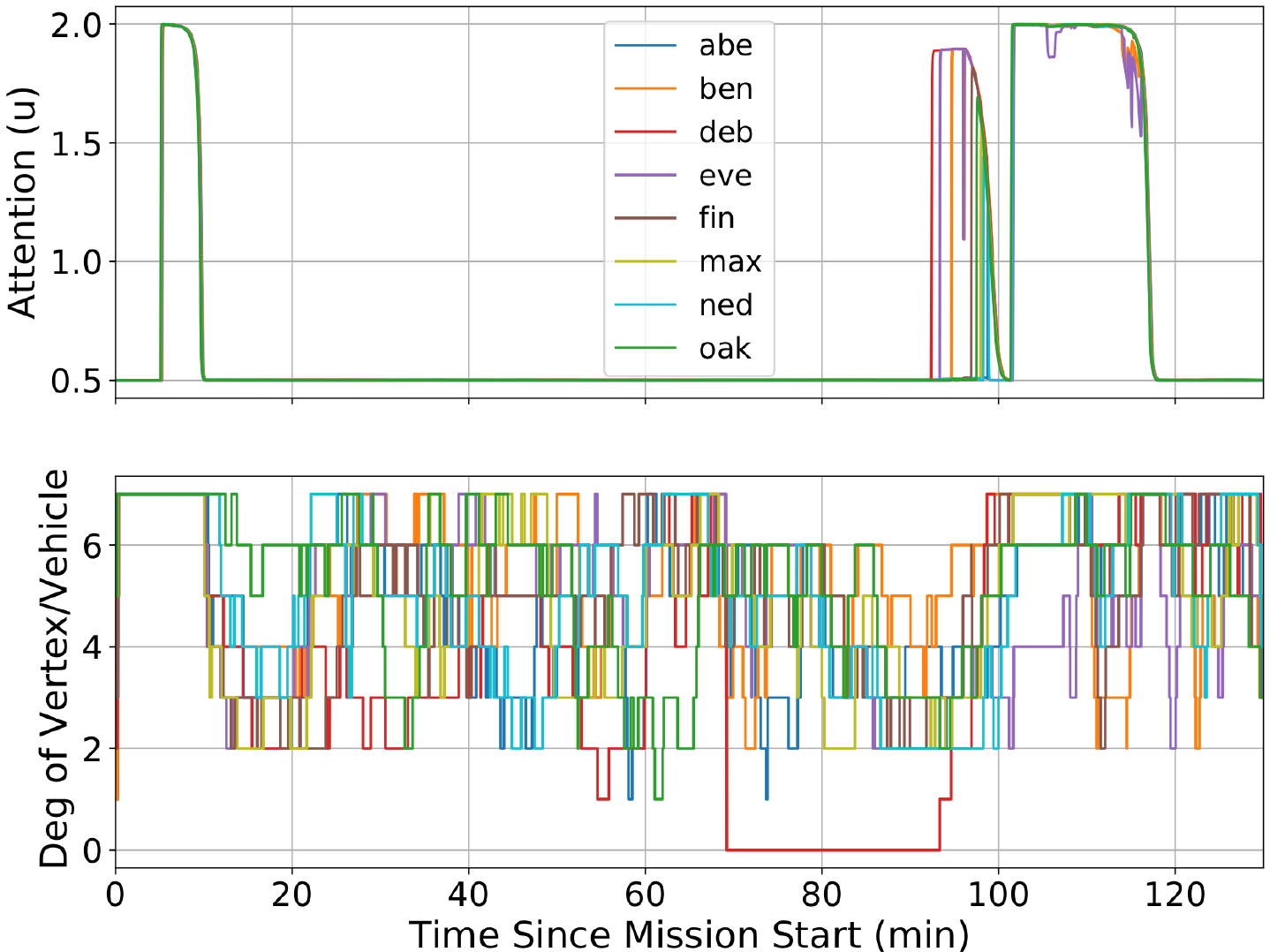}
    \caption{ Magnitude of attention (top) and degree of connection (bottom) for each vehicle throughout the 2 hour field mission.  By design vehicles had high attention during or after migration events (top). The number of neighbors within communication range fluctuated throughout the mission (bottom).}
    \label{fig:atten_deg}
    \vspace{-5mm}
\end{figure}

Approximately 70 minutes into the mission, the vehicle Deb unexpectedly stopped communicating, and because a dead-man switch was implemented in software,  Deb stopped the mission. The vehicle was manually driven back to the MIT sailing pavilion where it was revived and returned to the group about 25 minutes later.  This event can be seen in the bottom plot of Figure \ref{fig:atten_deg}, which shows the degree of connectivity for all vehicles - including the period when Deb maintained a degree of 0. 

This unplanned event demonstrated the robustness of our decentralized approach.  No additional commands were sent to the fleet while we recovered Deb, and the remaining 7 vehicles continued to dynamically form coalitions using the decentralized implementation of the Opinion Manager Engine.  Furthermore, the bottom plot on Figure \ref{fig:atten_deg} shows that during the mission every vehicle was periodically disconnected from others.  No single vehicle maintained a fully connected graph during the entire mission. 

\section{CONCLUSIONS AND FUTURE WORK}
In this paper we reported a new combination of group choice with individual decisions (GCID) and provided preliminary evaluation of its effectiveness using both Monte-Carlo simulation trails and a field demonstration. This approach models the choice of agents in a group as a  dynamical system of opinions, followed by a decision to optimize over multiple active behaviors using interval programming (IvP).  Results from randomized simulation studies suggest this new method offers improved performance over static coalitions.   In our field experiments we demonstrate collective behavior even with time-varying network topology and agent dropouts.   Future work includes determining bounds on performance of the system as a function of the local utility functions (\ref{eq:group_utility}), completing a more simulated comparison studies against other methods, and fielding other MRS with a different combination of vehicle types.

\section*{ACKNOWLEDGMENTS}
We would like to acknowledge all members of the MIT Pavlab who participated in field testing: Mikala Molina, Mark Franklin, Karan Mahesh, Raymond Turrisi, Kevin Becker, and Spring Lin.

\addtolength{\textheight}{-8.5cm}   % This command serves to balance the column lengths

%%%%%%%%%%%%%%%%%%%%%%%%%%%%%%%%%%%%%%%%%%%%%%%%%%%%%%%%%%%%%%%%%%%%%%%%%%%%%%%%

%%%%%%%%%%%%%%%%%%%%%%%%%%%%%%%%%%%%%%%%%%%%%%%%%%%%%%%%%%%%%%%%%%%%%%%%%%%%%%%%

%%%%%%%%%%%%%%%%%%%%%%%%%%%%%%%%%%%%%%%%%%%%%%%%%%%%%%%%%%%%%%%%%%%%%%%%%%%%%%%%

%%%%%%%%%%%%%%%%%%%%%%%%%%%%%%%%%%%%%%%%%%%%%%%%%%%%%%%%%%%%%%%%%%%%%%%%%%%%%%%%

\bibliographystyle{IEEEtran}
\bibliography{IEEEabrv,refs}

\begin{thebibliography}{10}
\providecommand{\url}[1]{#1}
\csname url@rmstyle\endcsname
\providecommand{\newblock}{\relax}
\providecommand{\bibinfo}[2]{#2}
\providecommand\BIBentrySTDinterwordspacing{\spaceskip=0pt\relax}
\providecommand\BIBentryALTinterwordstretchfactor{4}
\providecommand\BIBentryALTinterwordspacing{\spaceskip=\fontdimen2\font plus
\BIBentryALTinterwordstretchfactor\fontdimen3\font minus
  \fontdimen4\font\relax}
\providecommand\BIBforeignlanguage[2]{{%
\expandafter\ifx\csname l@#1\endcsname\relax
\typeout{** WARNING: IEEEtran.bst: No hyphenation pattern has been}%
\typeout{** loaded for the language `#1'. Using the pattern for}%
\typeout{** the default language instead.}%
\else
\language=\csname l@#1\endcsname
\fi
#2}}

\bibitem{Bizyaeva2023OD_TAC}
A.~Bizyaeva, A.~Franci, and N.~E. Leonard, ``Nonlinear opinion dynamics with
  tunable sensitivity,'' \emph{IEEE Transactions on Automatic Control},
  vol.~68, no.~3, pp. 1415--1430, 2023.

\bibitem{franci2021analysis}
A.~Franci, A.~Bizyaeva, S.~Park, and N.~E. Leonard, ``Analysis and control of
  agreement and disagreement opinion cascades,'' \emph{Swarm Intelligence},
  vol.~15, no. 1-2, pp. 47--82, 2021.

\bibitem{Benjamin2010MOOS}
M.~R. Benjamin, H.~Schmidt, P.~M. Newman, and J.~J. Leonard, ``Nested autonomy
  for unmanned marine vehicles with moos-ivp,'' \emph{Journal of Field
  Robotics}, vol.~27, no.~6, pp. 834--875, November/December 2010.

\bibitem{rizk2019cooperative}
Y.~Rizk, M.~Awad, and E.~W. Tunstel, ``Cooperative heterogeneous multi-robot
  systems: A survey,'' \emph{ACM Computing Surveys (CSUR)}, vol.~52, no.~2, pp.
  1--31, 2019.

\bibitem{verma2021multi}
J.~K. Verma and V.~Ranga, ``Multi-robot coordination analysis, taxonomy,
  challenges and future scope,'' \emph{Journal of intelligent \& robotic
  systems}, vol. 102, pp. 1--36, 2021.

\bibitem{Leonard2022PublicGoods}
N.~E. Leonard and S.~A. Levin, ``Collective intelligence as a public good,''
  \emph{Collective Intelligence}, vol.~1, no.~1, p. 26339137221083293, 2022.

\bibitem{Gershfeld2023RAL}
N.~Gershfeld, T.~M. Paine, and M.~R. Benjamin, ``Adaptive and collaborative
  bathymetric channel-finding approach for multiple autonomous marine
  vehicles,'' \emph{IEEE Robotics and Automation Letters}, vol.~8, no.~7, pp.
  4028--4035, 2023.

\bibitem{Park2021ICRA}
S.~Park, Y.~D. Zhong, and N.~E. Leonard, ``Multi-robot task allocation games in
  dynamically changing environments,'' in \emph{2021 IEEE International
  Conference on Robotics and Automation (ICRA)}, 2021, pp. 8678--8684.

\bibitem{Leonard2007IEEE}
N.~E. Leonard, D.~A. Paley, F.~Lekien, R.~Sepulchre, D.~M. Fratantoni, and
  R.~E. Davis, ``Collective motion, sensor networks, and ocean sampling,''
  \emph{Proceedings of the IEEE}, vol.~95, no.~1, pp. 48--74, 2007.

\bibitem{Kemna2017ICRA}
S.~Kemna, J.~G. Rogers, C.~Nieto-Granda, S.~Young, and G.~S. Sukhatme,
  ``Multi-robot coordination through dynamic voronoi partitioning for
  informative adaptive sampling in communication-constrained environments,'' in
  \emph{2017 IEEE International Conference on Robotics and Automation (ICRA)},
  2017, pp. 2124--2130.

\bibitem{Benjamin2021Oceans}
M.~R. Benjamin, T.~Paine, and S.~Randeni, ``Autonomy algorithms for stable
  dynamic linear convoying of autonomous marine vehicles,'' in \emph{OCEANS
  2021: San Diego – Porto}, 2021, pp. 1--10.

\bibitem{REN200RAS}
W.~Ren and N.~Sorensen, ``Distributed coordination architecture for multi-robot
  formation control,'' \emph{Robotics and Autonomous Systems}, vol.~56, no.~4,
  pp. 324--333, 2008.

\bibitem{Rubenstein2014Science}
M.~Rubenstein, A.~Cornejo, and R.~Nagpal, ``Programmable self-assembly in a
  thousand-robot swarm,'' \emph{Science}, vol. 345, no. 6198, pp. 795--799,
  2014.

\bibitem{couzin2009collective}
I.~D. Couzin, ``Collective cognition in animal groups,'' \emph{Trends in
  cognitive sciences}, vol.~13, no.~1, pp. 36--43, 2009.

\bibitem{Luc2009CBBA}
L.~Brunet, H.-L. Choi, and J.~How, ``Consensus-based decentralized auctions for
  robust task allocation,'' \emph{IEEE Trans. Robot.}, vol.~25, no.~4, pp.
  912--926, 2009.

\bibitem{Leahy2022ScRATCHes}
K.~Leahy, Z.~Serlin, C.-I. Vasile, A.~Schoer, A.~M. Jones, R.~Tron, and
  C.~Belta, ``Scalable and robust algorithms for task-based coordination from
  high-level specifications ({S}c{RATCH}e{S}),'' \emph{IEEE Transactions on
  Robotics}, vol.~38, no.~4, pp. 2516--2535, 2022.

\bibitem{liu2023catlnet}
W.~Liu, K.~Leahy, Z.~Serlin, and C.~Belta, ``Catlnet: Learning communication
  and coordination policies from catl+ specifications,'' in \emph{Learning for
  Dynamics and Control Conference}.\hskip 1em plus 0.5em minus 0.4em\relax
  PMLR, 2023, pp. 705--717.

\bibitem{tian2009multi}
Y.-T. Tian, M.~Yang, X.-Y. Qi, and Y.-M. Yang, ``Multi-robot task allocation
  for fire-disaster response based on reinforcement learning,'' in \emph{2009
  International Conference on Machine Learning and Cybernetics}, vol.~4.\hskip
  1em plus 0.5em minus 0.4em\relax IEEE, 2009, pp. 2312--2317.

\bibitem{Zheng2002MCTS}
H.~Zheng, J.~Guo, X.~Xie, and P.~Yan, ``A distributed coalition formation
  method of heterogeneous uav swarm in unknown dynamic environment,''
  \emph{Journal of Astronautics}, vol.~43, no.~2, pp. 189--197, 2022.

\bibitem{Karl2018MCTS}
K.~Kurzer, C.~Zhou, and J.~Marius~Zöllner, ``Decentralized cooperative
  planning for automated vehicles with hierarchical monte carlo tree search,''
  in \emph{2018 IEEE Intelligent Vehicles Symposium (IV)}, 2018, pp. 529--536.

\bibitem{Daneshvaramoli2020ComMCTS}
M.~Daneshvaramoli, M.~S. Kiarostami, S.~K. Monfared, H.~Karisani,
  K.~Dehghannayeri, D.~Rahmati, and S.~Gorgin, ``Decentralized
  communication-less multi-agent task assignment with cooperative monte-carlo
  tree search,'' in \emph{2020 6th International Conference on Control,
  Automation and Robotics (ICCAR)}, 2020, pp. 612--616.

\bibitem{benjamin2004interval}
M.~R. Benjamin, ``The interval programming model for multi-objective decision
  making,'' Computer Science and Artificial Intelligence Laboratory, MIT.
  Cambridge, MA, Tech. Rep. AIM-2004-021, 2004.

\bibitem{benjamin2010UserMan}
------, ``Moos-ivp autonomy tools users manual, mit computer science and
  artificial intelligence lab,'' Computer Science and Artificial Intelligence
  Laboratory, MIT. Cambridge, MA, Tech. Rep. MIT-CSAIL-TR-2010-039, 2010.

\bibitem{rome2021sensor}
M.~Rome, R.~E. Beighley, and T.~Faber, ``Sensor-based detection of algal blooms
  for public health advisories and long-term monitoring,'' \emph{Science of The
  Total Environment}, vol. 767, p. 144984, 2021.

\bibitem{evans2022practical}
N.~Evans, ``A practical search with voronoi distributed autonomous marine
  swarms,'' Master's thesis, Massachusetts Institute of Technology and Woods
  Hole Oceanographic Institution, 2022.

\bibitem{russell2010artificial}
S.~J. Russell, \emph{Artificial intelligence a modern approach}.\hskip 1em plus
  0.5em minus 0.4em\relax Pearson Education, Inc., 2010.

\bibitem{Becker2024Thesis}
K.~Becker, ``Development of a distributed simulation cluster for moos-ivp,''
  Master's thesis, Massachusetts Institute of Technology, 2024, under review.

\end{thebibliography}

\end{document}